\documentclass[preprint,3p,times,twocolumn]{elsarticle}

\usepackage{amssymb}
\usepackage{amsmath}
\usepackage{multicol}
\usepackage{multirow}
\usepackage{float}
\usepackage[colorlinks,bookmarksopen,bookmarksnumbered,citecolor=blue, linkcolor=blue, urlcolor=blue]{hyperref}

\journal{Neural Networks}

\let\today\relax
\makeatletter
\def\ps@pprintTitle{%
	\let\@oddhead\@empty
	\let\@evenhead\@empty
	\def\@oddfoot{\footnotesize\itshape
		{Submitted preprint} \hfill\today}%
	\let\@evenfoot\@oddfoot
}
\makeatother

\begin{document}

\begin{frontmatter}



\title{Combining Aggregated Attention and Transformer Architecture for Accurate and Efficient Performance of Spiking Neural Networks}

\author[label1]{Hangming Zhang} %
\author[label2,label3]{Alexander Sboev} 
\author[label2,label3]{Roman Rybka} 
\author[label1]{Qiang Yu\corref{cor1}}
\affiliation[label1]{organization={College of Intelligence and Computing},
            addressline={Tianjin University}, 
            city={Tianjin},
            postcode={300354}, 
            country={China}}

\affiliation[label2]{organization={National Research Center Kurchatov Institute},
            city={Moscow},
            postcode={123182},
            country={Russia}}

\affiliation[label3]{organization={National Research Nuclear University MEPhI},
            city={Moscow},
            postcode={115409},
            country={Russia}}
\cortext[cor1]{Corresponding author. E-mail address: yuqiang@tju.edu.cn}



\begin{abstract}
Spiking Neural Networks (SNNs), which simulate the spiking behavior of biological neurons, have attracted significant attention in recent years due to their distinctive low-power characteristics. Meanwhile, Transformer models, known for their powerful self-attention mechanisms and parallel processing capabilities, have demonstrated exceptional performance across various domains, including natural language processing and computer vision. Despite the significant advantages of both SNNs and Transformers, directly combining the low-power benefits of SNNs with the high performance of Transformers remains challenging. Specifically, while the sparse computing mode of SNNs contributes to reduced energy consumption, traditional attention mechanisms depend on dense matrix computations and complex softmax operations. This reliance poses significant challenges for effective execution in low-power scenarios. Traditional methods often struggle to maintain or enhance model performance while striving to reduce energy consumption. Given the tremendous success of Transformers in deep learning, it is a necessary step to explore the integration of SNNs and Transformers to harness the strengths of both. In this paper, we propose a novel model architecture, Spike Aggregation Transformer (SAFormer), that integrates the low-power characteristics of SNNs with the high-performance advantages of Transformer models. The core contribution of SAFormer lies in the design of the Spike Aggregated Self-Attention (SASA) mechanism, which significantly simplifies the computation process by calculating attention weights using only the spike matrices query and key, thereby effectively reducing energy consumption. Additionally, we introduce a Depthwise Convolution Module (DWC) to enhance the feature extraction capabilities, further improving overall accuracy. We evaluated SAFormer on the CIFAR-10, CIFAR-100, DVS128-Gesture, and CIFAR10-DVS datasets and demonstrated that SAFormer outperforms state-of-the-art SNNs in both accuracy and energy consumption, highlighting its significant advantages in low-power and high-performance computing.
\end{abstract}





\end{frontmatter}



\section{Introduction}
Spiking Neural Networks (SNNs) are widely considered as the third-generation neural network model due to their sparsity, low power consumption, and biological plausibility, which have attracted significant academic attention. Researchers have leveraged established architectures within artificial neural networks to construct SNNs, such as Recurrent Spiking Neural Networks\cite{zhang2019spike}, Spiking ResNet\cite{hu2021spiking}, Spiking Graph Neural Networks \cite{yin2024dynamic}, and Spiking GPT\cite{zhu2023spikegpt}. These efforts aim to enhance model performance through tailored architectural designs optimized for spiking computations. However, current SNNs are constrained by the use of sparse matrices, which limit their ability to capture and represent intricate patterns and relationships in the data. Due to the sparsity of spike-based computations, these matrices often lack the richness required for encoding complex, high-dimensional information. To address these challenges, recent research\cite{zhou2023spikformer, yao2024spike, zhou2023spikingformer} has integrated the Spiking Self-Attention (SSA) mechanism into SNNs, enabling them to learn and selectively focus on critical information
within inputs, facilitating operation in neuromorphic environments and achieving state-of-the-art performance
on static datasets.

The three inputs of SSA are query, key, and value matrices, all represented as the form of spikes. Similar to the Vanilla Self-Attention mechanism\cite{vaswani2017attention}, the attention matrix is computed by multiplying the query matrix with the transposed key matrix. The value matrix is then used to aggregate the relevant information, where each element is weighted by the attention scores to produce the final output. A notable advantage of SSA is its elimination of the softmax normalization step required in Vanilla Self-Attention mechanisms \cite{vaswani2017attention}. In the original architecture, softmax ensures that attention weights sum to 1, enforcing a normalized distribution to balance feature contributions. However, despite this simplification in attention computation, SSA still faces a challenge in terms of efficiency, particularly due to its quadratic complexity. This quadratic complexity arises from the need to evaluate interactions between each sequence element, which poses a challenge to the low-power efficiency in SNNs.

Some researchers have initiated investigations into reducing the computational complexity and power consumption of SSA, primarily categorized into two approaches. The first approach focuses on refining self-attention algorithms. These algorithms aim to design a linearly complex self-attention mechanism as an alternative to conventional SSA. For instance, Spike-Driven Self-Attention (SDSA) mechanism\cite{yao2024spike} replaces the conventional matrix multiplication of the query and key with the hadamard product. The approach relies solely on masking and addition operations, eliminating the need for any multiplication, thereby significantly reducing computational complexity. However, SDSA fails to address the issue of insufficient feature diversity caused by the sparsity of spike matrices, making it difficult for SNNs to concentrate on critical information in the input, which can ultimately reduce the overall accuracy. Such performance reduction is not a direct consequence of using the hadamard product itself but rather the lack of diversity in the feature representations. To address this limitation, a spike-based Q-K attention mechanism\cite{zhou2024qkformer} was proposed, which effectively captures the importance of token or channel dimensions using binary vectors with linear complexity. Nevertheless, the Q-K attention mechanism requires integration with SSA, which involves multiplication operations to ensure high performance. This combination, unfortunately, leads to an increase in the overall computational complexity of SNNs. In contrast, the Masked Spiking Transformer\cite{wang2023masked} (MST) was introduced as an approach to convert artificial neural networks to SNNs. MST restricts attention to local windows and facilitates inter-window information interaction to reduce computational demands while enhancing performance. However, MST typically requires more time steps to achieve the desired accuracy. Moreover, MST still maintains a quadratic time complexity, which increases computational time and power consumption. The second approach involves parameter-free linear transformations aimed at eliminating matrix multiplication to reduce computational complexity and power consumption. Wang et al.\cite{wang2023attention}  demonstrated that replacing the self-attention mechanism with fourier and wavelet transforms facilitates spike sequence integration, reducing quadratic time complexity to log-linear complexity. This approach achieves higher efficiency in large-scale tasks, but the lack of learnable parameters entails certain compromises in accuracy.

In this paper, we introduce a Spike Aggregation Transformer, a neural network framework designed to achieve high performance while minimizing power consumption, leveraging our proposed Spike Aggregated Self-Attention (SASA) mechanism. SASA offers the advantages of feature diversity and linear time complexity, encompassing three significant advancements. 

First, standard self-attention mechanisms typically depend on the value matrix to compute the final output representation. In contrast, SASA eliminates the value matrix and instead capitalizes on the diverse features generated by the key matrix, which contain rich information stemming from the sparsity and event-driven characteristics of SNNs. Diverging from conventional methods, we directly employ attention weights to modulate these features, generating the final output representation. This method simplifies the model structure while preserving critical information necessary for effective attention. 

Second, SASA is optimized for the redundancy and sparsity characteristic of global attention mechanisms. The computation of attention weights depends on the matrix multiplication of the query and key, which are typically sparse and high-dimensional. Consequently, the resulting attention weights exhibit significant sparsity and redundancy. To mitigate this issue, we utilize a down-sampling technique to derive low-dimensional aggregate matrices from the query and key. By applying the hadamard product to compute attention weights based on these aggregate matrices, we effectively reduce computational costs and circumvent the limitations imposed by sparsity. Furthermore, the down-sampling process reduces the sparsity of the spike matrix, resulting in a more expressive attention map. 

Third, similar to generalized linear attention mechanisms, SASA faces challenges with diversity of feature representations despite its efficiency improvements. To address this issue, we incorporate a Depthwise Convolution (DWC) module to enhance the diversity of features in SASA. DWC captures local features of the input data effectively by applying convolution operations separately to each input channel. This approach improves the effectiveness of feature extraction and further enriches the capability of the model to perceive diverse features.

In summary, our primary objective is to improve the efficiency and accuracy of SNNs while reducing computational complexity. Notably, although the down-sampling method we employed does not incorporate learnable parameters, the performance achieved surpasses that of traditional SSA methods. While down-sampling techniques involving learnable parameters are generally considered more effective, they may not always be suitable for SNNs with sparse and event-driven characteristics. Additionally, SASA demonstrates improved computational efficiency with a favorable linear computational complexity. Finally, we developed a SNNs based on SASA and conduct experiments on 4 challenging datasets.

The main contributions of this work are summarized as follows:
\begin{itemize}
\item We propose a novel SASA mechanism, which avoids operations involving the value matrix by leveraging diverse features derived from the key matrix, effectively reducing energy consumption. By focusing on key information through aggregate matrices, SASA improves the expressiveness of the attention map, thereby enhancing overall model performance.
\item We developed a high-performance, low-power Spike Aggregation Transformer framework based on the SASA. Since the aggregate matrices can be designed to be significantly smaller, the network offers an innovative solution for efficient learning on resource-constrained devices.
\item Extensive experiments demonstrate that the proposed architecture outperforms or matches state-of-the-art SNNs in terms of accuracy on both static and neuromorphic datasets, while achieving the lowest power consumption.
\end{itemize}
\section{Related Works}
\subsection{Transformer Architecture in SNNs}
In recent years, the incorporation of Transformer architectures into SNNs has attracted attention as a promising direction for enhancing both computational efficiency and performance. The initial development of the Transformer for natural language processing tasks, which utilized the attention mechanism to compute relationships between sequence elements, has proven beneficial in many areas of deep learning\cite{vaswani2017attention}. The Vision Transformer\cite{dosovitskiy2020vit} extended this architecture to computer vision tasks, demonstrating that Transformer models could perform competitively on image classification tasks. The simplicity, effectiveness, and scalability of Vision Transformers have stimulated numerous research in the field of computer vision. \cite{chen2023cf,dehghani2023scaling,lin2021fq,shi2023top,wang2023riformer,wang2023closer,liu2023efficientvit}. Recently, the integration of Transformer architecture into SNNs has opened new avenues for SNNs development. For instance, the Spikeformer model integrated spiking neurons into the feed-forward layers of a Transformer, combining spiking and non-spiking operations to create a hybrid system that retains energy efficiency advantages \cite{li2022spikeformer}. To address the energy inefficiency of the softmax operation in traditional attention mechanisms, the SSA mechanism was introduced\cite{zhou2023spikformer}. The sparse operations of SSA, devoid of multiplication, minimize computational energy consumption while maintaining high performance. However, a key limitation of existing Transformer-based SNNs models\cite{zhou2023spikformer,fang2021deep} lies in their reliance on non-spiking residual connections, which are incompatible with neuromorphic hardware designed to support fully spiking computations. In response to this challenge, Zhou et al.\cite{zhou2023spikingformer} developed a fully spiking Transformer model that eliminates non-spiking computations, resulting in a significant 57.34\% reduction in energy consumption. Recent studies \cite{zhou2023enhancing} identified inaccurate gradient backpropagation stemming from a suboptimal design of the downsampling module within the Transformer architecture. They proposed ConvBN-MaxPooling-LIF to optimize the downsampling process in SNNs, theoretically demonstrating that this method effectively rectifies gradient inaccuracies during backpropagation. These developments underscore the potential of Transformer-based architectures in SNNs, offering pathways to achieving high performance while maintaining the energy-efficient characteristics essential for neuromorphic computing.
\begin{figure*}[t]
  \centering
  \includegraphics[width=6.5in]{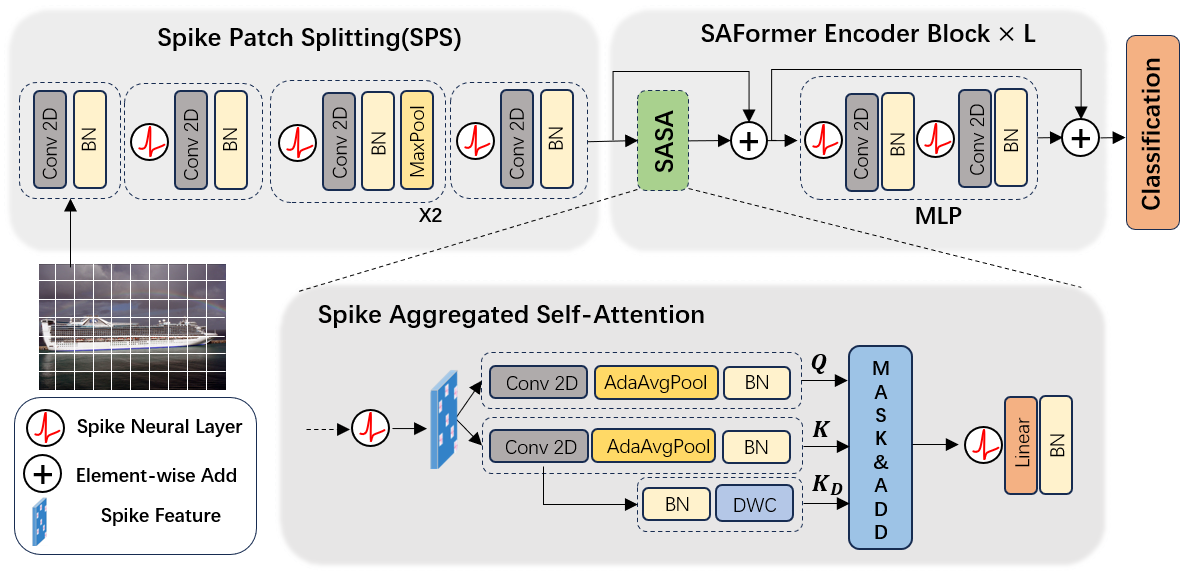}
  \caption{The overall architecture of our proposed method integrates a Spike Patch Splitting (SPS) module, multiple SAFormer encoder blocks, and a classification layer. SPS divides the image into multiple patches and then extracts multi-scale features from each patch. Next, stacking multiple Encoder Blocks facilitates efficient feature encoding, ultimately generating the prediction output through the classification layer. We have reconfigured the downsampling module and residual connection strategy to optimize the network structure, ensuring the flow of information in binary sparse matrix form throughout the neural network. The Encoder Blocks serve as core building units, combining the SASA mechanism with a multi-layer perceptron (MLP) module.}
  \label{model}
\end{figure*}
\subsection{Self-Attention Mechanisms in SNNs}
Attention mechanisms have proven crucial in enhancing neural networks by filtering out irrelevant data. In SNNs, the attention mechanism emulates how the human brain processes information. Rather than processing all data, it selectively emphasizes critical features, thereby enhancing information processing efficiency. Yao et al. \cite{yao2021temporal} were among the first to implement the attention mechanism in SNNs, aiming to filter out irrelevant spike sequences. However, the model primarily assesses the importance of information in the temporal dimension, assigning equal attention weights to all information within the same time step. Similarly, self-attention mechanisms confined to spatial domains have been proposed\cite{zhou2023spikformer}, neglecting temporal interactions across different time steps. To address these limitations, researchers developed spatiotemporal attention mechanisms\cite{zhu2024tcja} for SNNs, enabling the computation of feature dependencies across both time and space domains, and further implements asynchronous event processing\cite{wang2023spatial}. Further advancements include multi-dimensional attention mechanisms \cite{yao2023attention} that surpass models limited to one-dimensional information, as well as models \cite{yu2022stsc} inspired by biological synapses to enhance spatiotemporal receptive fields. In recent work, various attention mechanisms, such as denoising spike Transformers\cite{xu2023dista} and space-channel-temporal fused attention modules\cite{cai2023spatial}, have been introduced to help SNNs effectively handle complex spatiotemporal data. Despite these developments, existing mechanisms still face challenges due to the limitations of sparse matrices in expressing complex, high-dimensional data structures. 

To address this problem, we propose the SASA mechanism. By employing down-sampling techniques on the input data, we can generate aggregate matrices with reduced sparsity. The attention matrices computed from the aggregate matrices exhibit superior expressiveness. Furthermore, the aggregate matrices are significantly smaller in size compared to the input matrices, providing an effective solution for handling large-scale datasets. SASA reduces the number of computations and memory space while enhancing model efficiency and performance.

\section{Spike Aggregation Transformer}
We developed a Spike Aggregation Transformer (SAFormer) framework, which integrates the proposed SASA mechanism with the Transformer architecture within SNNs. Initially, we provide an overview of the spiking neuron model, followed by a detailed presentation of the SAFormer framework and its key components.
\subsection{Neuronal Model}
The spiking neuron model, which provides a simplified representation of biological neurons \cite{izhikevich2003simple,hodgkin1952quantitative}, encodes neuron activity as binary signals, where each spike indicates a neuronal firing event. In this study, we adopt the Leaky Integrate-and-Fire (LIF) neuron model as the fundamental unit. The dynamics of a LIF neuron can be formulated as follows:
\begin{equation}
H[t] = V[t-1] + \frac{1}{\tau} \left( X[t] - (V[t-1] - V_{\mathrm{reset}}) \right)
\end{equation}
\begin{equation}
S[t] = \Theta(H[t] - V_{\mathrm{th}})
\end{equation}
\begin{equation}
V[t] = H[t] (1 - S[t]) + V_{\mathrm{reset}} S[t]
\end{equation}
Here, $t$ denotes the time step, $V[t-1]$ represents the membrane potential at the previous time step, and $\tau$ denotes the membrane time constant, which governs the rate of information decay. A larger $\tau$ results in slower information decay. $X[t]$ denotes the current input at time step $t$, and $H[t]$ represents the output generated by $X[t]$ and $V[t-1]$. When $H[t]$ surpasses the threshold $V_{\mathrm{th}}$, the neuron fires a spike $S[t]$. The Heaviside step function $\Theta(v)$ is defined such that $\Theta(v)=1$ if $v\ge0$, and $\Theta(v)=0$ otherwise. $V[t]$ represents the membrane potential after a spike event:$V[t]=H[t]$ if no spike is generated, indicating insufficient information to trigger a spike; otherwise, $V[t]= V_{\mathrm{reset}}$ after a spike, with the membrane potential returning to the reset value $V_{\mathrm{reset}}$ in preparation for the next spike.

\subsection{Network Architecture}
Figure \ref{model} illustrates the overall architecture of our SAFormer, comprising three main components: spike patch splitting (SPS), $L$ encoder blocks, and a linear classification head. Previous studies \cite{zhou2023spikformer} have shown that layer normalization is not effective for SNNs. Therefore, we opted to use batch normalization instead. Given a 2D image sequence $I\in\mathbb{R}^{T\times C\times H\times W}$, where $T$ denotes the time steps (for static image datasets, $T$ is the number of repetitions), $C$, $H$, and $W$ represent the channel and spatial dimensions of the image sequence, the SPS module (see Section \ref{SPS}) generates processed patch block feature maps $U_0$, capturing both local details and semantic information: 
\begin{align} U_0 &= \mathrm{SPS}(I), \quad U_0 \in \mathbb{R}^{T \times N \times D} \end{align} 
Here, $U_0$ denotes the output feature sequence, where $N$ represents the sequence length, and $D$ represents the number of elements per patch sample. $U_0$ is then fed into the SAFormer encoder block, which integrates SASA mechanism (see Section \ref{SASA}) and a MLP module. Residual connection is employed for membrane potential in both the SASA and the MLP. Formulated as:
\begin{equation} 
U_{l}=\mathrm{SASA}(U_{l-1})+U_{l-1},U_{l}\in\mathbb{R}^{T\times N\times D},l=1\ldots L
\end{equation} 
\begin{equation} 
S_{l}=\mathrm{MLP}(U_{l})+U_{l},S_{l}\in\mathbb{R}^{T\times N\times D},l=1\ldots L 
\end{equation} 
Here, $U_l$ and $S_l$ represent the outputs of the $l$-th layer of the SASA mechanism and the encoder block, respectively. The features outputted by the last encoder undergo global average pooling (GAP) to yield $D$-dimensional channels, followed by classification via a fully-connected-layer classification head (CH) to produce prediction $Y$, yielding:
\begin{equation} Y = \mathrm{CH}(\mathrm{GAP}(S_{L})) \end{equation} 

\subsection{Spike Patch Splitting}\label{SPS}
As depicted in Figure \ref{model}, the primary function of the SPS module is to preprocess the input image by segmenting it into multiple uniformly sized blocks, referred to as patches. Each patch maintains consistent dimensions, with the aim of enhancing spatial continuity. 

Following the recent architectural design\cite{zhou2023spikformer,yao2024spike}, Our SPS module employs five convolutional layers to process the input image. Initially, the output channels of the first four convolutional layers are set to $D/8, D/4, D/2$, and $D$. By expanding the number of channels, the network is able to capture a broader range of features at each layer, thereby facilitating the learning of both basic and intricate patterns in the input data. The fifth convolutional layer integrates these hierarchical features, yielding a feature map with $D$-dimensional channels. These feature maps encompass local details like edges and textures, providing a rich representation of features for the subsequent encoder blocks. Our convolutional layer design optimizes down-sampling operations for enhanced performance. Specifically, maxpooling is applied after the third and fourth convolutional layers. This decision was made to reduce data dimensionality and computational load, while preserving important spatial details. Applying maxpooling too early could result in excessive information loss, especially given the sparse and binary of spike-based transmission in SNNs. Following the architectural design of Spikformer\cite{zhou2023spikformer}, pooling is delayed to these later layers to manage network complexity while preserving information integrity. To further enhance performance and address gradient backpropagation inaccuracies\cite{zhou2023enhancing}, we integrate a spike neural layer immediately after the pooling operation. This addition effectively addresses gradient backpropagation inaccuracies, thereby enhancing network training efficiency and overall performance.

\subsection{Spike Aggregated Self-Attention Mechanism}\label{SASA}
The SASA mechanism is a vital component of the network architecture and will be discussed in conjunction with the SSA\cite{zhou2023spikformer} mechanism. Given the feature sequence $X$, the SSA mechanism utilizes three essential matrices: query($Q_{\mathrm{S}}$), key($K_{\mathrm{S}}$), and value($V_{\mathrm{S}}$), which are derived from learnable linear transformations $W_{\mathrm{Q}},W_{\mathrm{K}},W_{\mathrm{V}}\in\mathbb{R}^{D\times D}$ and $X$:
\begin{align}
Q_{\mathrm{S}}=\mathrm{SN}\left(\mathrm{BN}\big(XW_\mathrm{Q}\big)\right)\notag\\
K_{\mathrm{S}}=\mathrm{SN}\left(\mathrm{BN}\big(XW_\mathrm{K}\big)\right)\\
V_{\mathrm{S}}=\mathrm{SN}\left(\mathrm{BN}\big(XW_\mathrm{V}\big)\right)\notag
\end{align}
The output of SSA is computed as:
\begin{align}
&\mathrm{SSA}^{\prime}(Q_{\mathrm{S}},K_{\mathrm{S}},V_{\mathrm{S}})=\mathrm{SN}(Q_{\mathrm{S}}K_{\mathrm{S}}^{T}V_{\mathrm{S}}* c)
\end{align}
\begin{equation}
\mathrm{SSA}(Q_{\mathrm{S}},  K_{\mathrm{S}},V_{\mathrm{S}})=\mathrm{SN}(\mathrm{BN}(\mathrm{Linear}(\mathrm{SSA}^{\prime}(Q_{\mathrm{S}},K_{\mathrm{S}},V_{\mathrm{S}}))))
\end{equation}

Here, $c$ denotes the scaling factor, $\mathrm{SN}$ represents the spike neural layer. An important enhancement of the SSA mechanism is its ability to compute without the softmax function, leveraging the non-negative attention map it produces, which eliminates the need for additional normalization. This design simplifies computation while preserving the original characteristics of the attention mechanism, thereby enhancing the capture of significant information from the input sequence. However, it is important to acknowledge that SSA encounters difficulties in terms of computational complexity. The computation time grows quadratically with sequence length because SSA evaluates pairwise comparisons across all sequence elements when calculating attention weights. Consequently, processing longer sequences substantially increases computational demands, which restricts the applicability of SSA in contexts involving large-scale data or real-time processing.

To address these challenges, we propose the SASA mechanism. As shown at the bottom of Figure \ref{model}, our SASA primarily employs two spike-form matrices: query ($Q$) and key ($K$). The matrices are generated from learnable linear transformations $W_\mathrm{Q}$ and $W_\mathrm{K}$, which efficiently capture essential information from input data for subsequent attention calculations. SASA capitalizes on the sparsity of spike-based matrices, which reduces computational complexity and enhances model robustness. However, this sparsity can also impact the accuracy of attention calculations. To mitigate this, we introduce an aggregation function $AG(\cdot)$ to perform down-sampling on the floating-point matrices $Q_\mathrm{F}$ and $K_\mathrm{F}$. This function aggregates sparse spike signals into denser and more stable feature representations, thereby enhancing the accuracy of attention calculations. In our approach, we employ adaptive average pooling, a widely recognized function in deep learning, as the aggregation function because of its flexibility in generating feature maps of various sizes based on input characteristics. By adjusting the pooling window sizes and strides, adaptive average pooling effectively captures multi-scale feature information, enhancing performance in handling complex data scenarios\cite{ding2024novel}. The formulation proceeds as follows:
\begin{equation}
Q_{\mathrm{F}}=XW_{\mathrm{Q}},K_{\mathrm{F}}=XW_{\mathrm{K}},Q_{\mathrm{F}},K_{\mathrm{F}}\in\mathbb{R}^{T\times N\times D}  
\end{equation}
\begin{equation}
Q=\mathrm{SN}_{}\big(\mathrm{BN}(AG(Q_{\mathrm{F}}))\big),Q\in\mathbb{R}^{T\times n\times D}
\end{equation}
\begin{equation}
    K=\mathrm{SN}_{}\big(\mathrm{BN}(AG(K_{\mathrm{F}}))\big),K\in\mathbb{R}^{T\times n\times D}
\end{equation}
Here, $n$ denotes the sequence length after aggregation, which is significantly reduced compared to $N$. Subsequently, $Q$ and $K$ are used to compute the attention map:
\begin{equation}
    \mathrm{SASA}^{\prime}(Q,K)=\mathrm{SN}\left(\mathrm{SUM}_{\mathrm{c}}(Q\otimes K)\right)
\end{equation}
Here, $\otimes$ denotes the hadamard product, and $\mathrm{SUM}_\mathrm{c}(\cdot)$ denotes column-wise summation. Notably, SASA diverges from traditional attention mechanisms by omitting the value ($V$) matrix. This divergence is due to the sparse and event-driven of SNNs, which allow for efficient information processing through selective neuron activation and computation synchronization in response to specific events. In this context, the $V$ matrix aggregates information from all inputs. However, it becomes largely redundant because only a limited subset of neurons engages in processing at any moment. In contrast, traditional attention mechanisms rely on $V$ for storing and transmitting feature information derived from input data. This information is then weighted and summed based on the attention matrix to highlight important parts of the input data. This streamlined approach, achieved by eliminating the $V$ matrix, reduces computational overhead and improves the efficiency of processing extensive and complex datasets. However, this simplified structure limits feature diversity. To address this issue, we utilize the DWC module to extract diverse features from the $K$ matrix and add them to the attention map. In contrast to standard convolution operations, the DWC module takes advantage of depthwise separable convolution, which significantly reduces computational complexity while extracting richer features. This approach enhances feature diversity and significantly amplifies the overall expressive capacity of the model. It demonstrates greater adaptability and robustness\cite{han2023agent,han2023flatten}, especially when handling complex and multi-dimensional data.
\begin{align}
    &\hat{K_\mathrm{D}}=\mathrm{SN}\big(\mathrm{BN}(K_\mathrm{F})\big)&\hat{K_{\mathrm{D}}}\in\mathbb{R}^{T\times N\times D}
\end{align}
\begin{align}
    &K_\mathrm{D}= \mathrm{DWC(\hat{K_\mathrm{D}})}=\mathrm{SN}(\mathrm{BN}(\hat{K_\mathrm{D}}_{*d}W_\mathrm{D}))&K_{\mathrm{D}}\in\mathbb{R}^{T\times N\times D}
\end{align}
\begin{equation}
\mathrm{SASA}(Q,K)=\mathrm{BN}\left(\mathrm{Linear}\left(\mathrm{SN}\left(K_\mathrm{D}\oplus \mathrm{SASA}^{\prime}(Q,K)\right)\right)\right) 
\end{equation}
Here, $\hat{K_\mathrm{D}}$ represents the pre-processed version of $K_\mathrm{D}$, while $\hat{K_\mathrm{D}}_{*d}W_\mathrm{D}$ refers to the depthwise convolution of $\hat{K_\mathrm{D}}$ with the weight matrix $W_\mathrm{D}$. $K_\mathrm{D}$ denotes the enhanced feature diversity, and $\oplus$ indicates element-wise addition. This formulation ensures SASA maintains comprehensive feature representation while optimizing computational efficiency.
\section{Experiments}
In this section, we evaluate the performance of our SAFormer on the static image datasets\cite{krizhevsky2009learning}, and explore its effectiveness on neuromorphic datasets\cite{li2017cifar10,amir2017low}. Our evaluation includes detailed analysis of metrics such as classification accuracy and computational efficiency. We also assess the robustness and generalizability of our SAFormer through a series of ablation studies that dissect the core architectural components.
\subsection{Results on Static Datasets Classification}
\begin{table*}[h]
\centering
\caption{Experimental results on CIFAR-10 and CIFAR-100. SAFormer-L-D represents a SAFormer model with L encoder blocks and D feature embedding dimensions.}
\label{cifar_experiment_table}
\renewcommand\arraystretch{1.3}{
\begin{tabular}{lccccc}
\hline
\multirow{2}{*}{Methods} &  \multirow{2}{*}{Architecture} &\multicolumn{2}{c}{CIFAR-10} & \multicolumn{2}{c}{CIFAR-100} \\ \cline{3-6} 
                         &                              & T      & Acc     & T       & Acc     \\ \hline
Hybrid training\cite{rathi2020enabling}  &    VGG-11  & 125 & 92.22 & 125 & 67.87 \\
SALT\cite{kim2021optimizing}  &    VGG-9/11  & 40 & 87.10 & 40 & 59.11 \\
PLIF\cite{fang2021incorporating}   &     VGG-8      & 8     & 93.50    & -            & -       \\
STBP\cite{wu2018spatio} & CIFARNet  & 12 & 89.83 & - & - \\
STBP NeuNorm\cite{wu2019direct} & CIFARNet  & 12 & 90.53 & - & - \\
TSSL-BP\cite{zhang2020temporal} & CIFARNet  & 5 & 91.41 & - & - \\
Dspike\cite{li2021differentiable}   &       ResNet-18                                      & 6           & 94.30    & 6        & 74.20    \\
tdBN\cite{zheng2021going} &      ResNet-19                                            & 6        & 93.20    & 4          & 70.86       \\
TET\cite{deng2022temporal} & ResNet-19  & 4 & 94.44 & 4 & 74.47 \\
DIET-SNN\cite{rathi2021diet}     &   ResNet-20                                    & 5           & 92.70    & 5        & 69.67   \\
MS-ResNet\cite{hu2024advancing}   &      MS-ResNet-21/18                                    & 16           & 88.96   & 6             & 76.41       \\
SEW-ResNet\cite{fang2021deep}   &    SEW-ResNet-21/18                                     & 16          & 90.02   & 6              & 77.37       \\

DSR\cite{meng2022training}     &       PreAct-ResNet-18                                       & 20           & 95.40    & 20         & 78.50    \\
Spikformer\cite{zhou2023spikformer}  &      Spiking Transformer-4-384                                     & 4              & 95.50    & 4            & 78.20    \\
S-Transformer\cite{yao2024spike}   &        Spiking Transformer-4-384                                      & 4           & 95.60    & 4             & 78.40    \\ 
 \hline
ANN    &   ResNet-19                                 & 1           & 94.97   & 1         & 75.40    \\ \hline
\textbf{This Work}    &    SAFormer-4-384                                         &4  &\textbf{95.80} & 4  & \textbf{79.07}\\ \hline
\end{tabular}}
\end{table*}

In this experiment, we followed the experimental setup of Spikformer \cite{zhou2023spikformer} and S-Transformer \cite{yao2024spike} to evaluate the performance of our SAFormer. We trained the model for 400 epochs to ensure sufficient learning and optimization time. A batch size of 64 was used to balance training efficiency and memory usage. For optimization, we employed AdamW\cite{loshchilov2017fixing}, a variant of the Adam optimizer that incorporates weight decay, which is particularly effective for handling large-scale datasets and complex models, often leading to superior performance\cite{liu2021swin}. Our SAFormer included 4 encoder blocks with 384 feature embedding dimensions. Each block in the model possesses learning capabilities, and their sequential stacking enhances the ability of model to represent intricate and abstract features\cite{vaswani2017attention,dosovitskiy2020vit}. 
\begin{figure}[htbp]
  \centering
  \includegraphics[width=3.2in]{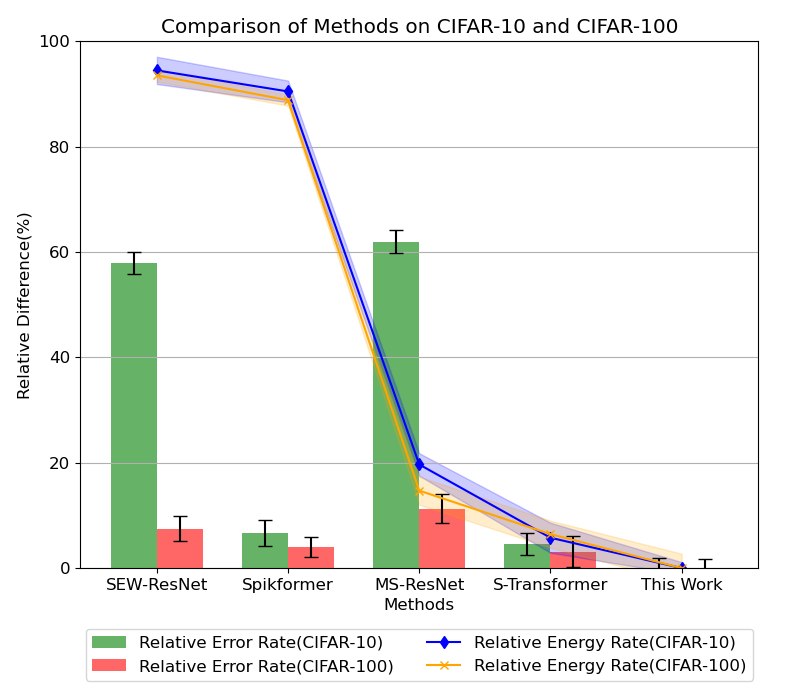}
  \caption{Performance comparison of different methods on the CIFAR-10 and CIFAR-100 datasets. Both error rates and energy efficiency of our model relative to each method are shown to highlight the significance of our model. The averages and standard deviations are calculated over three independent runs.}
  \label{cifar_experiments_figure}
\end{figure} 

To assess the energy consumption of SAFormer, it is essential to first compute the number of synaptic operations involved:
\begin{equation}
    \mathrm{SOP}(l)=fr\times T\times\mathrm{FLOPs}(l)
\end{equation}
Let $l$ represent a block in SAFormer, where $fr$ denotes the firing rate of the input spike sequence, and $T$ is the simulation time step of the spiking neurons. FLOPs($l$) represents the floating point operations of block $l$, which corresponds to the number of multiply-and-accumulate (MAC) operations. In SNNs, a spike is treated as an event that triggers an accumulation process. These accumulation processes are analogous to accumulate (AC) operations, where the accumulated value is updated in response to incoming spikes\cite{gerstner2014neuronal}. Based on the assumption that both MAC and AC operations are implemented on 45nm hardware, following the energy model in \cite{horowitz20141}, where the energy per MAC operation ($E_{MAC}$) is 4.6 pJ and the energy per AC operation ($E_{AC}$) is 0.9 pJ, the theoretical energy consumption of SAFormer is computed as follows:
\begin{equation}\label{energy_equation}
    E_{\mathrm{SAFormer}}=E_{MAC}\times\mathrm{FL}_{\mathrm{Conv}}^{1} + E_{AC}\times\mathrm{SP}
\end{equation}
\begin{equation}
\mathrm{SP}=\sum_{p=2}^{P}\mathrm{SOP}_{\text{Conv}}^{p}+\sum_{m=1}^{M}\mathrm{SOP}_{\text{FC}}^{m}+\sum_{l=1}^{L}\mathrm{SOP}_{\text{SASA}}^{l}
\end{equation}
Here, $\mathrm{FL}_{\mathrm{Conv}}^{1}$ refers to the number of operations consumed by the first layer responsible for encoding static RGB images into spike form, while $\mathrm{SP}$ denotes the number of spike-based operations for the remaining layers, which include the $p$ SNN convolutional layers, the $m$ SNN fully connected layers, and the $l$ SASA layers. To quantify the difference between models, we employ the relative difference metric, which is defined as:
\begin{equation}
    RD_i = \frac{|I_{our} - I_{i}|}{ I_i}
\end{equation}
where $RD_i$ represents the relative difference between the models. $I_i$ represents the value of a specific metric (such as accuracy rate, error rate or energy consumption) for the current model, and $I_{our}$ denotes the corresponding value for our model. 
\begin{table*}[t]
\centering
\caption{Experimental results on CIFAR10-DVS and DVS128-Gesture.}
\label{dvs_experiment_table}
\renewcommand\arraystretch{1.3}{
\begin{tabular}{lccccc}
\hline
\multirow{2}{*}{Methods} & \multirow{2}{*}{Architecture} & \multicolumn{2}{c}{CIFAR10-DVS} & \multicolumn{2}{c}{DVS128-Gesture}  \\ \cline{3-6} 
            &                                           & T             & Acc             & T  & Acc                                \\ \hline
LIAF-Net\cite{wu2021liaf} & -  & 10 & 70.4 & 60 & 97.6 \\
TA-SNN\cite{yao2021temporal} & -  & 10 & 72.0 & 60 & 98.6 \\
Rollout\cite{kugele2020efficient} & DenseNet  & 48 & 66.8 & 240 & 97.2 \\
DECOLLE\cite{kaiser2020synaptic} & 3-layer CNN  & - & - & 500 & 95.5 \\
SEW-ResNet\cite{fang2021deep} & 7B-Net  & 16 & 74.4 & 16 & 97.9 \\
PLIF\cite{fang2021incorporating}  &      VGG-6/7                  & 20            & 74.8            & 20 & 97.6                             \\
DSR\cite{meng2022training}       &       VGG-11              & 10            & 77.3            & -  & -                                 \\
SALT\cite{kim2021optimizing} & VGG-16  & 20 & 67.1 & - & - \\

Dspike\cite{li2021differentiable}  &     ResNet-18                                          & 10            & 75.4            & -  & -                                               \\
tdBN\cite{zheng2021going}     &    ResNet-19                                         & 10            & 67.8            & 40 & 96.9                            \\
MS-ResNet\cite{hu2024advancing} &  MS-ResNet-20  & - & 75.6 & - & - \\

Spikformer\cite{zhou2023spikformer}    &    Spiking Transformer-2-256                                  & 16            & 80.9            & 16 & 98.3                                               \\
S-Transformer\cite{yao2024spike}    &      Spiking Transformer-2-256                 & 16            & 80.0              & 16 & \textbf{99.3}                      \\ \hline

\textbf{This Work}     &    SAFormer-2-256                   & 16            &  \textbf{81.3}               & 16 & 98.3                            \\
\hline
\end{tabular}}
\end{table*}

The experimental results on the CIFAR dataset clearly demonstrate that our SAFormer model enhances accuracy while maintaining low energy consumption. The detailed experimental results are provided in Table \ref{cifar_experiment_table} and visualized in Figure \ref{cifar_experiments_figure}, highlighting the relative differences in evaluate metrics between models. For CIFAR-10 dataset, our SAFormer model achieved an accuracy of 95.8\% , with an energy consumption of only 0.49 mJ, calculated according to Equation \ref{energy_equation}. This represents a substantial improvement over the currently popular ResNet architecture \cite{hu2024advancing}. Moreover, our model achieved a 19.67\% reduction compared to MS-ResNet\cite{hu2024advancing} in energy consumption, underscoring its suitability for energy-efficient applications. Comparisons with SEW-ResNet \cite{fang2021deep} further demonstrated the superiority of SAFormer, showing a 5.78\% increase in accuracy and a 94.48\% reduction in energy consumption. These findings underscore the exceptional energy efficiency and performance enhancements of our SAFormer. In addition to comparisons with ResNet-based models, we evaluated SAFormer against Transformer architecture models. While the accuracy improvement over Spikformer \cite{zhou2023spikformer} was modest (0.3\%), our SAFormer excelled in energy efficiency with a substantial 90.49\% reduction in energy consumption. Similarly, compared to S-Transformer \cite{yao2024spike}, SAFormer achieved a slight 0.2\% accuracy improvement coupled with a notable 5.8\% reduction in energy consumption. These results further establish the advantageous balance of our SAFormer between energy efficiency and accuracy.

On the CIFAR-100 dataset, SAFormer achieved high accuracy while maintaining minimal energy consumption, as illustrated in Table \ref{cifar_experiment_table} and Figure \ref{cifar_experiments_figure}. With an energy consumption of 0.58 mJ, calculated according to Equation \ref{energy_equation}. Our model achieved an accuracy of 79.07\%, surpassing other advanced models. Specifically, compared to DSR\cite{meng2022training} and Dspike\cite{li2021differentiable} methods employing ResNet architecture, SAFormer exhibited improvements in accuracy by 0.57\% and 4.87\%, respectively. These results underscore the effectiveness of our SAFormer in handling complex datasets with diverse categories in image classification tasks. Similarly, although the accuracy improvements over Spikeformer \cite{zhou2023spikformer} and S-Transformer \cite{yao2024spike} are modest (0.87\% and 0.67\%, respectively), our SAFormer significantly excelled in energy efficiency. Compared to Spikformer, SAFormer achieved a 88.82\% reduction in energy consumption, and compared to S-Transformer, it achieved a 6.5\% reduction.

\subsection{Results on Neuromorphic Datasets Classification}
We compare our method with state-of-the-art approaches on the CIFAR10-DVS and DVS128-Gesture datasets, strictly following the experimental protocols outlined in Spikformer \cite{zhou2023spikformer} and S-Transformer \cite{yao2024spike}. Our model configuration includes 2 encoder blocks and 256 feature embedding dimensions, designed to capture essential information within the data. The time step parameter is set to 16, striking a balance between model complexity and computational efficiency. The training process spans 300 epochs for CIFAR10-DVS and 400 epochs for DVS128-Gesture.

\begin{figure}[t]
  \centering
  \includegraphics[width=3.2in]{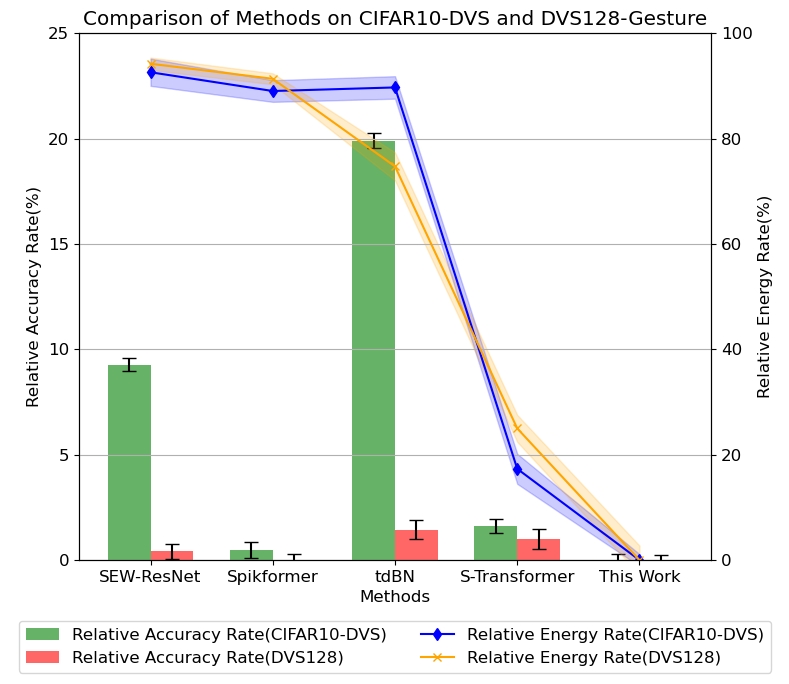}
  \caption{The performance comparison on the CIFAR10-DVS and DVS128-Gesture datasets underscores the advantages of our model in terms of both accuracy rates and energy efficiency relative to other methods. The averages and standard deviations are calculated over three independent runs.}
  \label{dvs_experiments_figure}
\end{figure}

The detailed results of the neuromorphic classification experiments are presented in Table \ref{dvs_experiment_table} and Figure \ref{dvs_experiments_figure}. On the CIFAR10-DVS dataset, our model achieved an accuracy of 81.3\% with an energy consumption of only 1.67 mJ. This marks a significant improvement over ResNet-based methods such as SEW-ResNet \cite{fang2021deep} and tdBN \cite{zheng2021going}, with accuracy increases of 6.9\% and 13.5\% respectively, while energy consumption is reduced by 92.6\% and 89.7\%. Additionally, compared to Spikformer\cite{zhou2023spikformer} and S-Transformer\cite{yao2024spike} methods, our approach achieves accuracy improvements of 0.4\% and 1.3\% respectively, while reducing energy consumption by 89.0\% and 17.3\%. These results further validate the superior performance and effectiveness of our model in neuromorphic classification tasks.

On the DVS128-Gesture dataset, our model demonstrates outstanding performance, achieving an accuracy of  98.3\% with an energy consumption of 1.23 mJ. Compared to traditional methods like tdBN \cite{zheng2021going} and PLIF \cite{fang2021incorporating}, our model exhibited clear advantages. Specifically, compared to tdBN, our model improves accuracy by 1.4\% and reduces energy consumption by 74.7\%, highlighting its capability to discern subtle changes and complex patterns in the data. Similarly, compared to PLIF, our model achieves a 0.7\% increase in accuracy, further underscoring its robust performance in neuromorphic classification tasks. Additionally, our model matches the performance of state-of-the-art methods\cite{yao2024spike} while reducing energy consumption by 25\%, enhancing its effectiveness and competitiveness in the field.

\subsection{Ablation Study}
In this part, we conduct an ablation study on the DWC module and $AG$ function to analyze their specific impacts on model performance. We use the SDSA\cite{yao2024spike} as our benchmark, consisting of 2 encoder blocks with 512 embedding dimensions.

\begin{figure}[t]
  \centering
  \includegraphics[width=2.0in]{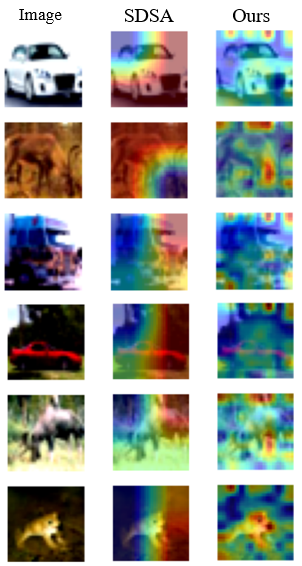}
  \caption{We randomly selected six images (left column) from the CIFAR-10 dataset and visualized the attention maps generated by SDSA (middle column) and our SASA (right colume). The color intensity of the attention maps indicates the degree of attention assigned to each region, with darker colors representing higher attention.}
  \label{attention_map}
\end{figure}

\begin{table}[h]
\centering
\caption{Ablation studies on SAFormer-2-512. D and A represent DWC module and $AG$ function respectively.  When both components are excluded, the primary difference between our method and SDSA lies in the use of the $V$ matrix.}
\renewcommand\arraystretch{1.3}{
    \begin{tabular}{ccc}
    \hline
    Model                    & CIFAR-10 & CIFAR-100 \\ \hline
   SDSA\cite{yao2024spike} & 93.82   & 74.41    \\ \hline
    w\textbackslash{}o D   & 95.45(+1.63)        & 78.51(+4.10)         \\
    w\textbackslash{}o A   & 95.64(+1.82)        & 78.29(+3.88)         \\
    w\textbackslash{}o D+A   & 95.41(+1.59)        &78.18(+3.77)         \\ \hline
    This Work                 & 95.70(+1.88)        & 78.54(+4.13)         \\ \hline
    \end{tabular}}
    \label{ablation_study}
\end{table}
Table \ref{ablation_study} presents the performance on CIFAR-10 and CIFAR-100 datasets. Our experimental results show that using either the DWC module or the $AG$ function alone can achieve better performance than the SDSA method. We also present a visualization of the attention maps for SASA and SDSA to analyze the enhancements achieved by the SASA mechanism in focusing on key information, as shown in Figure \ref{attention_map}. The SDSA method shows a more dispersed attention distribution, whereas our SASA method utilizes an aggregation function to reduce sparsity and employs a DWC module to enhance feature diversity. These enhancements enable our SASA method to achieve more effective attention allocation.


Notably, the best performance of our model is observed when both the DWC module and $AG$ function are present, achieving 95.7\% on CIFAR-10 and 78.54\% on CIFAR-100. These results validate the effectiveness of the DWC module and $AG$ function in improving model accuracy on challenging complex tasks.

\subsection{Impact of Time Step on Performance}
\begin{figure}[t]
  \centering
  \includegraphics[width=2.8in]{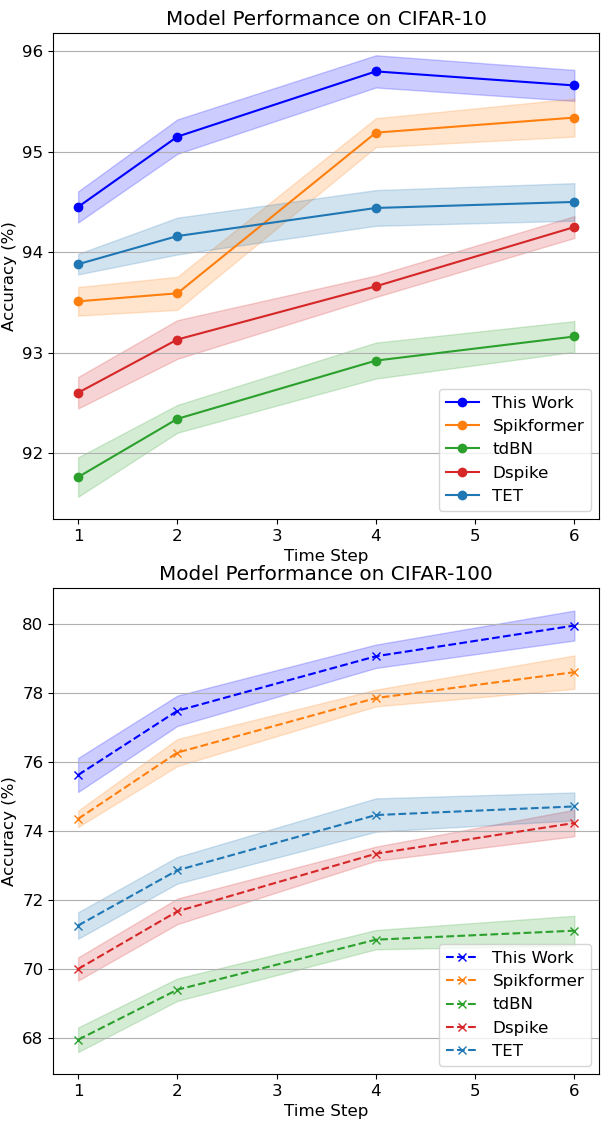}
  \caption{The performance of five different deep learning models over time on the standard image classification benchmark datasets CIFAR-10 and CIFAR-100.}
  \label{time_step}
\end{figure}

The experimental results for different simulation time steps are detailed in Figure \ref{time_step}. Our analysis reveals a clear trend: as the time step increases, the accuracy of the compared models significantly improves. Although our method experiences a marginal decrease in accuracy at $T=6$ compared to $T=4$ on the CIFAR-10 dataset, it still outperforms the other models, achieving the highest accuracy overall. The observed performance decrease can be attributed to the limited number of categories in the CIFAR-10 dataset, where the feature information extracted by our SAFormer is saturated at T=4. However, no accuracy decrease is observed on the CIFAR-100 dataset, which has a larger number of categories, thus providing that the performance decline is due to the smaller number of categories.

Notably, our SAFormer model achieves higher accuracy at $T=1$ than the TET\cite{deng2022temporal}, Dspike\cite{li2021differentiable}, and tdBN\cite{zheng2021going} models at $T=6$. This advantage is mainly attributed to the SAFormer model adopting our SASA mechanism, which can enhance the feature representation of the attention map, thereby reducing the simulation time step and improving the accuracy, which is potentially suitable for edge computing and other resource-constrained environments.

\subsection{Effects of Sequence Length After Aggregation}
\begin{figure}[t]
  \centering
  \includegraphics[width=3.15in]{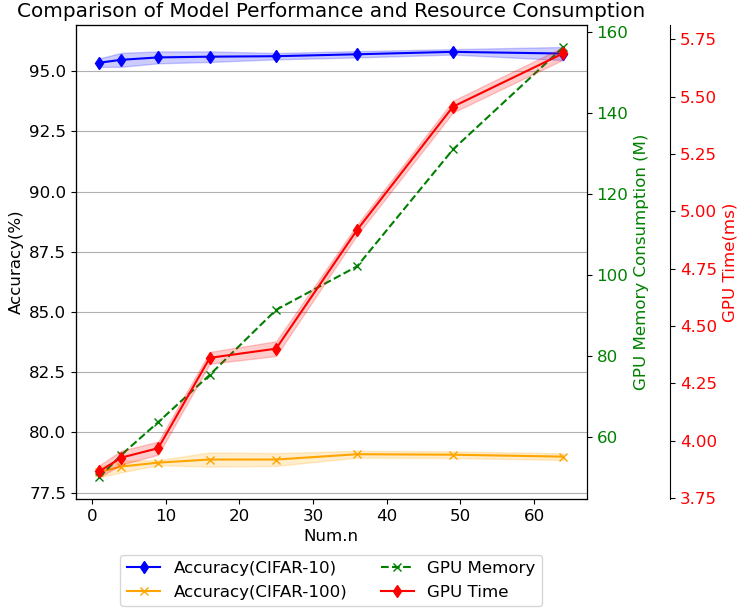}
  \caption{The impact of hyperparameter $n$ on model performance and computational resource usage. The graph depicts the trends in accuracy for the CIFAR-10 and CIFAR-100 datasets, as well as GPU memory consumption and GPU execution time.} 
  \label{hyperparameter_n}
\end{figure}
We focused on a hyperparameter $n$, which denotes the reduced sequence length after aggregation. Under the fixed condition of a time step of 4 and batch size of 64, we systematically evaluated its impact on model performance, GPU memory usage, and execution time. As illustrated in Figure \ref{hyperparameter_n}, this experimental analysis reinforces the rationale behind our model design and highlights its unique advantages.

It is noteworthy that even when $n$ is set to the minimum value of 1, our model exhibits performance comparable to current state-of-the-art SNNs\cite{yao2024spike}. This outcome indicates that our SASA mechanism is sufficiently robust in its simplest setup, capable of effectively capturing and processing critical information within the data. This finding not only validates the effectiveness of our model design but also provides a solid foundation for subsequent optimization efforts.

As the value of $n$ gradually increases, the performance of the model gradually saturates, while the changes in computing resource consumption become significant. Specifically, GPU memory usage rises substantially with increasing $n$, primarily due to the growth in internal parameters and computational complexity. Additionally, execution time extends considerably due to the increased volume of computations required for each iteration. 

\section{Conclusion}
We develop SAFormer that combines the low power consumption characteristics of SNNs with the high-accuracy strength of transformer models. The core of SAFormer is our SASA mechanism. SASA effectively aggregates spike features, allowing the model to capture and utilize crucial information. Its linear complexity also ensures scalability to larger datasets and more complex tasks. 

Our experimental results on both static and neuromorphic datasets demonstrate that SAFormer not only achieves leading accuracy performance but also excels in energy consumption, underscoring its potential and value in practical applications. Further research will focus on refining the SASA mechanism to enhance its robustness and adaptability across diverse applications. Additionally, exploring more advanced normalization techniques and optimization algorithms could further enhance performance and efficiency. In summary, SAFormer represents an advancement in the development of energy-efficient neural networks, offering a promising direction for future research and practical applications in areas such as image classification, object detection, and beyond. By continuing to refine and expand upon this innovative approach, we aim to contribute to the broader goal of creating more efficient and effective artificial intelligence systems.

\appendix
\section{Additional Analysis and Discussions on the Efficiency of SASA}
\subsection{Computational Complexity Analysis}
\begin{table}[htbp]
\centering
\caption{Comparison of computational complexity of attention mechanisms. $n, N$ represents the number of Tokens, and $D$ represents the number of channels.}
\renewcommand\arraystretch{1.3}{
\begin{tabular}{ccc}
\hline
Methods    & Time & Space \\
\hline
VSA\cite{vaswani2017attention}        &  $\mathcal{O}(N^2D)$    &  $\mathcal{O}(N^2D+ND)$     \\
SSA\cite{zhou2023spikformer}        &  $\mathcal{O}(N^2D)$    &  $\mathcal{O}(N^2D+ND)$      \\
SDSA\cite{yao2024spike}       &  $\mathcal{O}(ND)$    &   $\mathcal{O}(ND)$    \\
\hline
SASA(Ours) &  $\mathcal{O}(nD)$    &    $\mathcal{O}(nD)$   \\
\hline
\end{tabular}}
\label{complexity}
\end{table}
As shown in Table \ref{complexity}, the time complexity of SASA reaches $O(nD)$. This significant advantage makes SASA more efficient when processing large-scale dataset. SASA successfully reduces time complexity by streamlining calculation steps and reducing matrix operations, allowing it to complete calculation tasks more quickly when processing large-scale dataset. It should be noted that the DWC module is integrated into the SASA mechanism to enhance feature diversity and is analyzed separately in the complexity assessment. Although convolution operations are generally nonlinear, the DWC module employs efficient implementations to reduce computational overhead, ensuring that its inclusion does not significantly affect the linear time complexity of the overall model. 

The space complexity of SASA reaches $O(nD)$, which is due to the storage space consumption of $Q$ and $K$ matrices during the calculation of our SASA mechanism. Since SASA does not use an additional $V$ matrix, it has an advantage in space occupancy over the traditional self-attention mechanism. Reduced space complexity enables SASA to utilize computing resources more efficiently, especially in memory-constrained environments.

\begin{table}[htbp]
\centering
\caption{Energy evaluation. $fr_i$ denote the spike firing rates in various spike matrices.}
\renewcommand\arraystretch{1.3}{
\begin{tabular}{ccc}
\hline
Methods & Operations & Energy  \\ \hline
\multirow{4}{*}{VSA\cite{vaswani2017attention}} &  $Q,K,V$                    &   $E_{MAC}*3ND^2$                   \\
                      &   $f(Q,K,V)$                   &      $E_{MAC}*2N^2D$                \\
                      &   Scale                   &      $E_{MAC}*N^2$                \\
                      &   Softmax                   &      $E_{MAC}*2N^2$                \\ \hline
\multirow{2}{*}{SSA\cite{zhou2023spikformer}} &  $Q,K,V$                    &   $E_{AC}*T*fr_1*3ND^2$                   \\
                      &   $f(Q,K,V)$                   &      $E_{AC}*T*fr_2*2N^2D$                \\ \hline
\multirow{2}{*}{SDSA\cite{yao2024spike}} &  $Q,K,V$                    &   $E_{AC}*T*fr_3*3ND^2$                   \\
                      &   $f(Q,K,V)$                   &      $E_{AC}*T*fr_4*ND$                \\ \hline
\multirow{2}{*}{SASA(Ours)} &      $Q,K$                &    $E_{AC}*T*fr_5*2ND^2$                  \\
                      & $f(Q,K)$ & $E_{AC}*T*fr_6*nD$ \\
\hline
\end{tabular}}
\label{energy_consumption}
\end{table}
\subsection{Energy Efficiency Analysis}
In our SASA mechanism, all matrix elements are represented in binary form, which transforms the original complex linear multiplication into a more efficient sparse addition operation. This conversion not only simplifies the computational process but also significantly reduces power consumption, as demonstrated in Table \ref{energy_consumption}. In contrast to VSA \cite{vaswani2017attention}, SSA \cite{zhou2023spikformer}, and SDSA \cite{yao2024spike}, our SASA mechanism relies solely on the $Q$ and $K$ matrices for efficient attention calculation, eliminating the need for the $V$ matrix and complex scaling operations. As a result, the number of synaptic computations is reduced, leading to enhanced efficiency.


\section*{Acknowledgments}
Financial support by the National Natural Science Foundation of China (Grant 92370103 and 62176179) and the Xiaomi Foundation is gratefully acknowledged.





\bibliography{reference}
\bibliographystyle{unsrt}



\end{document}